\documentclass{article}

\PassOptionsToPackage{sort&compress}{natbib}
\usepackage[final]{corl_2026}

\usepackage{amsmath}
\usepackage{amssymb}
\usepackage{booktabs}
\usepackage{graphicx}
\usepackage{float}
\usepackage{tabularx}
\usepackage{wrapfig}
\usepackage{makecell}
\usepackage{multicol}
\usepackage{subcaption} %
\usepackage{xcolor}
\usepackage{hyperref}
\usepackage[capitalise]{cleveref}

\graphicspath{{./}{./figures/}}

\newcolumntype{x}[0]{>{\centering\arraybackslash}X}

\newcommand{\numTraj}{3{,}593}
\newcommand{\numCams}{20}

\title{AutoDex: An Automated Real-World System for Dexterous Grasping Data Collection}

\author{
  Mingi Choi$^{1}$ \quad Gunhee Kim$^{1}$ \quad Jisoo Kim$^{1}$ \quad Taeksoo Kim$^{1}$\\
  \textbf{Taeyun Ha$^{1}$ \quad Jongbin Lim$^{1}$ \quad Hanbyul Joo$^{1,2}$}\\[4pt]
  $^{1}$Seoul National University \qquad $^{2}$RLWRLD\\[2pt]
  {\small\texttt{\{willi19,gunhee2001,jlogkim,taeksu98,taeyun012,whdqls0534,hbjoo\}@snu.ac.kr}}
}

\hypersetup{
  pdftitle={AutoDex: An Automated Real-World System for Dexterous Grasping Data Collection},
  pdfauthor={Mingi Choi, Gunhee Kim, Jisoo Kim, Taeksoo Kim, Taeyun Ha, Jongbin Lim, Hanbyul Joo},
  pdfsubject={},
  pdfkeywords={Dexterous Grasping, Autonomous Data Collection, Sim-to-Real}
}

\begin{document}
\maketitle

\begin{abstract}
Scaling real-world dexterous-grasp data collection is limited by the sustained human effort required to operate the robot, verify grasp outcomes, and reset objects between trials. Simulation can generate grasp candidates at scale, but their success on hardware remains uncertain. We present \textbf{AutoDex}, a system that collects real-world dexterous-grasp data without continuous human operation. Given generated grasp candidates, AutoDex plans and executes grasps, labels their outcomes, and resets the object to continue collection across stable poses. Multi-view object tracking supports labeling under hand--object occlusion, while residual-torque collision monitoring and pose-directed reset enable the loop to operate without continuous human supervision. 
Using AutoDex, we collect over 3,500 grasp trials across 100 objects with Allegro and Inspire hands, recording both successes and failures together with synchronized visual observations and robot states. 
AutoDex reduces human effort by more than an order of magnitude, requiring an estimated 0.2 human-hours to collect 500 trajectories compared with 9.3 for teleoperation.
As AutoDex collects trials, it builds a library of physically validated grasps for subsequent execution. In new object placements and obstacle configurations, retrieval from this library achieves 76\% success, compared with 34\% for candidates screened only in simulation. Code and data are available at \url{https://snuvclab.github.io/AutoDex/}.
\end{abstract}

\keywords{Dexterous Grasping, Autonomous Data Collection, Sim-to-Real}
\suppressfloats[t]

\begin{figure}[t]
    \centering
    \includegraphics[width=\textwidth]{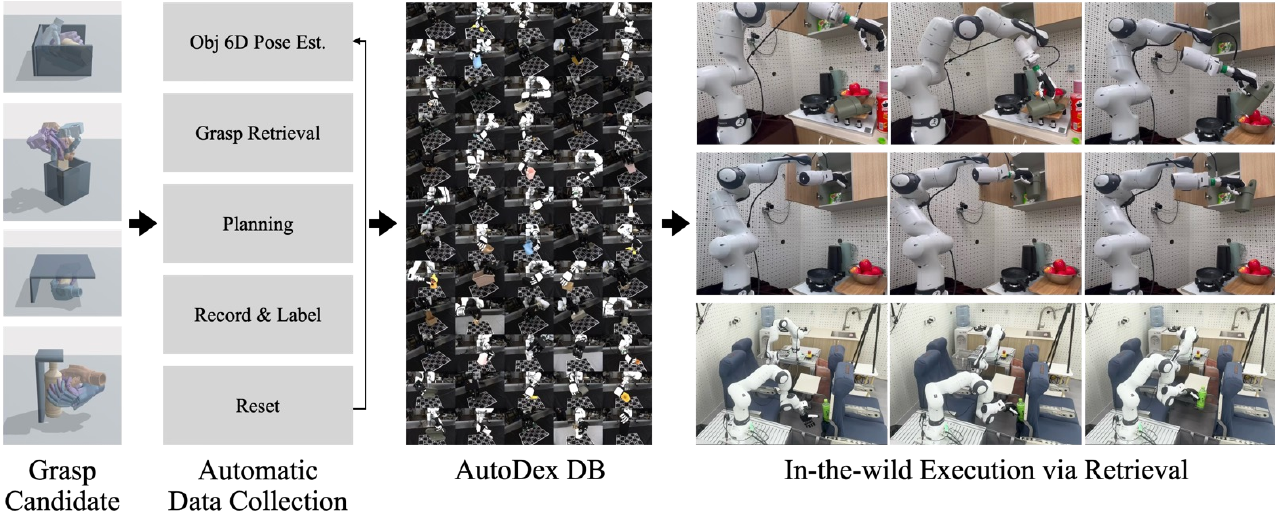}
    \vspace{-14pt}
    \caption{
    \textbf{The AutoDex pipeline.}
    AutoDex builds a database of physically labeled dexterous-grasp trials by executing generated candidates in a multi-camera workcell, labeling lift-and-hold success or failure, and resetting the object between trials. At deployment, downstream systems retrieve successful grasps, filter them for feasibility in the new scene, and execute the selected grasp.
    }
    \vspace{-16pt}
    \label{fig:teaser}
\end{figure}

\section{Introduction}
\label{sec:intro}

Learning robust dexterous grasping requires data that records not only proposed hand--object configurations, but also their physical outcomes on real hardware~\cite{liu2024realdex,wang2024dexcap,chen2022dextransfer,zhang2024dexgraspnet20learninggenerative}. Unlike parallel-jaw grasps, multi-finger grasps involve high-dimensional hand configurations and multiple simultaneous contacts, so small errors in pose, friction, compliance, or force distribution can determine whether an apparently plausible grasp succeeds or slips~\cite{bicchi2000grasping}. As a result, geometric feasibility alone is not a reliable proxy for grasp validity: a candidate may be collision-free and kinematically feasible while still failing under real contact dynamics. Dexterous grasp datasets should therefore include both successful and failed grasp attempts with ground-truth physical outcome labels across diverse objects, poses, materials, and scene constraints.

Existing collection paradigms for such data trade off scale against physical validity. Human teleoperation~\cite{liu2024realdex,wang2024dexcap,qin2023anyteleop} produces real contact outcomes, but it is slow, fatigue-limited, and biased by the operator's choices over approach directions and finger configurations. In contrast, optimization- and simulation-based grasp synthesis methods~\cite{wang2022dexgraspnet,chen2024bodex,turpin2023fast} can generate large sets of candidate hand--object configurations at low cost. Many aspects of candidate validity are model-based and largely geometric: with accurate object meshes, robot calibration, and collision models, one can check joint-limit satisfaction, IK solvability, rigid-body collision avoidance, and motion-planning feasibility reasonably reliably. However, these checks only establish that a candidate is geometrically executable. They do not certify whether the resulting multi-finger contact will stably support the object. Real grasp success depends on contact physics---friction, slip, compliance, finger-pad deformation, actuator behavior, and the distribution of contact forces---which are difficult to simulate accurately. Thus, a candidate can be generator-valid and planner-executable, yet fail in physical execution~\cite{huang2025dexgrasp}.

A natural solution is to combine scalable candidate synthesis and model-based feasibility checking with real-world verification: use synthesis and planning tools to propose executable candidate grasps, then test whether those candidates succeed under real physical interaction. However, hardware validation scales only if the full loop is automated. Any manual intervention in initial object pose estimation, execution-time tracking under hand--object occlusion, candidate selection, safety monitoring, success/failure labeling, or scene reset reintroduces the human bottleneck of teleoperation. The challenge is not merely grasp generation, but end-to-end autonomous data collection.

To this end, we present AutoDex, an automated real-world data-collection system that turns generated grasp candidates into physically labeled robot trials. Given an object, AutoDex obtains candidate grasps from a modular candidate generator, filters and selects executable candidates, executes selected grasps on a physical multi-finger robot hand, records synchronized robot-state and multi-view visual data with success/failure labels, and resets the object for the next trial. AutoDex realizes this automation through four system components: (1) dense \numCams-camera perception for initial object pose estimation and execution-time tracking under hand--object occlusion, (2) collision-monitored motion execution for safe unattended grasp attempts, (3) physical success/failure labeling from lift-and-hold trials, and (4) active object reset for continued testing across stable poses. Together, these components enable fully automated real-world dexterous-grasp data collection at scale.

Using AutoDex, we collect \numTraj{} automatically labeled real-world grasp trials with Allegro and Inspire hands across 100 diverse objects. In a 500-trajectory collection study, AutoDex requires 10.3\,h of wall time and 0.2\,h of human attention, compared with 28.4\,h and 9.3\,h, respectively, for teleoperation. Across scene types, grasps retrieved from the AutoDex-validated database achieve 76\% real-world success, compared with 34\% for grasps selected without real-world validation. Our contributions are threefold.
First, we introduce AutoDex, an end-to-end automated system for real-world execution and labeling of generated dexterous-grasp candidates.
Second, we collect a large real-world grasp-trial database containing both successes and failures, synchronized multi-view observations, and robot-state trajectories across 100 objects.
Third, we provide a system-level evaluation of the full collection loop, including throughput against teleoperation, real-world success against a model-screened baseline, autonomous reset across difficult pose transitions, and the effect of dense multi-view perception on collection reliability.

\section{Related Work}
\label{sec:related}

\textbf{Dexterous-grasp data collection and synthesis.}
Human demonstrations provide a source of dexterous manipulation data through robot teleoperation and hand motion capture. Teleoperation enables the collection of real robot grasping trajectories~\cite{liu2024realdex,lim2026hrdexdb,wang2025dexh2r,li2026bimanualgrasp}, with vision-based interfaces supporting different arm--hand systems~\cite{qin2023anyteleop,handa2020dexpilot,cheng2024opentelevision,ding2024bunnyvisionpro}. Other approaches capture human hand motions for imitation learning~\cite{wang2024dexcap,chen2022dextransfer,qin2022dexmv,shaw2022videodex,tao2025dexwild}. Hand-held grippers and wearable exoskeletons provide another way to record demonstrations with robot-compatible kinematics~\cite{chi2024umi,fang2024airexo,xu2025dexumi}, while DexUMI uses robot-hand inpainting to reduce the visual gap between human demonstrations and robot execution~\cite{xu2025dexumi}. Augmented- and virtual-reality interfaces also enable demonstration collection without physical robot hardware, either through teleoperation in simulation~\cite{park2025dart} or with virtual robot feedback rendered over real scenes~\cite{chen2025arcap}. These approaches make demonstration collection more accessible, but still require a person to perform each demonstration. These limitations motivate autonomous physical validation of simulation-generated multi-finger grasp candidates.

Grasp synthesis offers another route to scaling dexterous-grasp data. Existing methods use optimization~\cite{miller2004graspit,Liu_2022,wang2022dexgraspnet,turpin2022graspd,turpin2023fast,chen2024bodex}, grasp-type templates~\cite{chen2025dexonomy}, learned generative models~\cite{xu2023unidexgrasp,zhang2024dexgraspnet20learninggenerative,feng2025ffhflow}, and reinforcement learning~\cite{christen2022d,qin2022dexpoint,wan2023unidexgrasp++,zhang2025robustdexgrasp} to produce candidate hand configurations. By generating grasp candidates at scale and validating them in simulation, these methods construct large datasets for learning grasp synthesis and evaluation~\cite{wang2022dexgraspnet,xu2023unidexgrasp,chen2024bodex,lum2025getagrip,chen2025dexonomy,ye2025dex1b}. The resulting grasps can be paired with synthetic observations and grasp-quality labels to train grasp evaluators~\cite{mahler2017dexnet2,mousavian2019graspnet,lum2025getagrip}. However, the quality of a grasp under an optimization objective or a simulation model does not guarantee its success on real hardware~\cite{horak2019contact,rubert2019predicting,lelidec2024contact,lum2025getagrip}. AutoDex complements these synthetic datasets by executing generated grasp candidates on hardware and recording both successes and failures.

\textbf{Autonomous real-world data collection.}
Large-scale self-supervised grasping~\cite{pinto2016supersizing,gupta2018homes,levine2018handeye,kalashnikov2018qtopt} and multi-task reinforcement learning~\cite{kalashnikov2021mtopt} demonstrate how repeated robot interaction can support policy learning. Subsequent work improves policies autonomously through online adaptation~\cite{sharma2023selfimproving,zhou2024soar}, learning during deployment with human interventions~\cite{liu2023sirius}, or fleet-level data collection that combines autonomous execution with teleoperation~\cite{ahn2024autort}. Other systems address the practical requirements of long-running collection through learned reward prediction and autonomous reset~\cite{yang2023robofume}, while dexterous reinforcement learning refines grasping policies pretrained from demonstrations~\cite{huang2025dexgrasp}. These approaches show that autonomous interaction can scale robot data collection, but they primarily collect policy-driven experience, and many focus on parallel-jaw manipulation. They do not provide a general mechanism for executing and physically labeling large sets of simulation-generated multi-finger grasp candidates, where reliable success detection and pose-specific reset remain central challenges~\cite{eysenbach2018leave,zhu2020ingredients,gupta2021resetfree,du2023success}. AutoDex addresses this gap by combining autonomous execution with outcome measurement and pose-directed reset, thereby turning simulation-generated candidates into a reusable database of real-world successes and failures.

\section{Method}
\label{sec:method}

AutoDex is an automated system for collecting real-world dexterous-grasp trials with physical outcome labels. Given an object and generated candidate grasps, AutoDex estimates the object pose, filters and selects executable candidates, executes the selected grasp on a physical multi-finger hand, labels the trial as success or failure, records synchronized robot and visual data, and resets the object to continue collection, all without human intervention.

\subsection{Grasp Candidate Generation}
\label{sec:method:generator}
AutoDex begins by obtaining candidate grasps from a modular candidate generator. The generator takes an object model and a scene specification as input and returns a set of candidate grasps, each represented by a wrist pose in the object frame and a hand configuration. The scene specification describes the manipulation environment, such as a tabletop with surrounding obstacles.
The generator is responsible for proposing kinematically and geometrically feasible candidates, such as wrist poses and hand configurations that are collision-free with respect to the object and scene. It is not assumed to certify stability under real contact dynamics. AutoDex instead tests selected candidates through real-world execution and labels each trial by lift-and-hold success or failure.

In this work, we use BODex~\cite{chen2024bodex} as the candidate generator and instantiate the scene specification with three planar constraint primitives: wall, shelf, and box. These primitives cover common tabletop constraints in which one or more approach directions are blocked, while excluding fully enclosed configurations that are unreachable by the hand. Let \(\mathcal{P}\) denote the set of stable tabletop poses of the object. For each stable pose \(P_i \in \mathcal{P}\), we aggregate candidates generated over scene primitives into a pose-indexed set \(\mathcal{G}_{\mathrm{cand}}(P_i)\). At runtime, AutoDex considers candidates for the current stable pose, invokes reset when no feasible unattempted candidates remain at that pose, and continues across stable poses until no further feasible candidates are available or the target dataset size is reached. The leftmost column in \cref{fig:teaser} shows representative candidates.

\begin{figure*}[t]
\centering
\includegraphics[width=\textwidth]{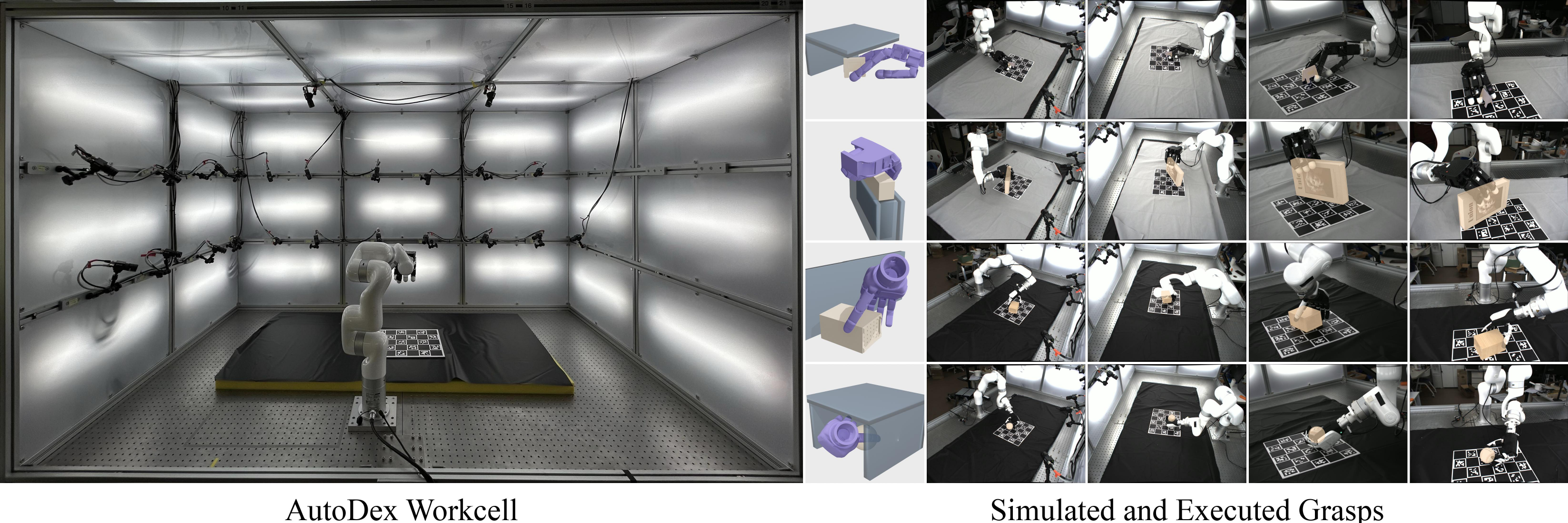}
\setlength{\abovecaptionskip}{2pt}
\caption{
\textbf{AutoDex workcell and execution examples.}
\textbf{Left:} A multi-camera workcell with a 6-DoF xArm, a swappable multi-finger hand, and \numCams{} synchronized RGB cameras.
\textbf{Middle and right:} Each row pairs a candidate grasp generated under a wall, shelf, or box scene constraint with its corresponding real-world execution, shown with synchronized views and overlaid tracked 6D object poses.
}
\label{fig:system}
\end{figure*}

\subsection{AutoDex Workcell}
\label{sec:method:system}
The AutoDex workcell is a calibrated robot-capture system designed to execute dexterous grasp trials without human intervention while recording synchronized visual and robot-state data. It consists of a 6-DoF xArm equipped with a swappable multi-finger hand, mounted inside a 1.5$\times$1.5$\times$2.5 m LED-lit capture cell with \numCams{} synchronized RGB cameras, as shown in \cref{fig:system}.
Each camera records at 30 FPS with 2048$\times$1536 resolution and is synchronized by sub-millisecond hardware triggers. The dense camera arrangement provides redundant views for object pose estimation during grasp execution, when the hand can severely occlude the object. Camera intrinsics are calibrated before mounting using a ChArUco board, and per-session extrinsics are recovered with COLMAP and hand-eye calibration, yielding sub-millimeter multi-view pose self-consistency in our \numCams-camera setting.
The robot controller operates at 120 Hz, streaming joint positions, velocities, torques, motor currents, and commanded targets from both the arm and hand. In our experiments, we use either a 16-DoF Allegro Hand or a 6-DoF Inspire Hand, attached through a standardized arm-hand mounting interface. A calibrated robot-camera timestamp offset aligns robot-state measurements with the multi-view RGB observations. 

Since AutoDex runs without human intervention, the workcell requires an automatic safety mechanism for unexpected contacts during execution. The robot arm's built-in collision detector was unreliable in our setting because grasped objects appear as persistent external loads and the mounted dexterous hand changes the end-effector inertia, both of which can trigger false positives. We therefore use a learned residual-torque monitor trained on collision-free motions of the deployed arm-hand assembly. Given joint position \(q\), joint velocity \(\dot q\), and motor-current-derived torque \(\tau_{\mathrm{motor}}\), the monitor predicts the nominal free-space torque with an MLP and computes the residual:
\[
    \hat{\tau}_{\theta} = \mathrm{MLP}_{\theta}(q,\dot q),
    \qquad
    \tau_{\mathrm{res}} = \hat{\tau}_{\theta} - \tau_{\mathrm{motor}} .
\]
A sustained residual above a preset safety threshold is treated as unexpected contact. To avoid unnecessary aborts, we enable the monitor only during contact-critical segments, such as downward motions near the tabletop or surrounding scene. When such contact is detected, execution is halted, and the robot recovers by planning a trajectory back to a predefined home configuration while monotonically increasing the end-effector height. Together, dense multi-view perception and residual-torque safety monitoring provide precise object-state estimation and execution safeguards needed for autonomous dataset collection without human supervision.

\subsection{Autonomous Data Collection Loop}
\label{sec:method:pipeline}

\begin{figure}[t]
\centering

\begin{minipage}[t]{0.56\linewidth}
\centering
\includegraphics[height=4.29cm]
{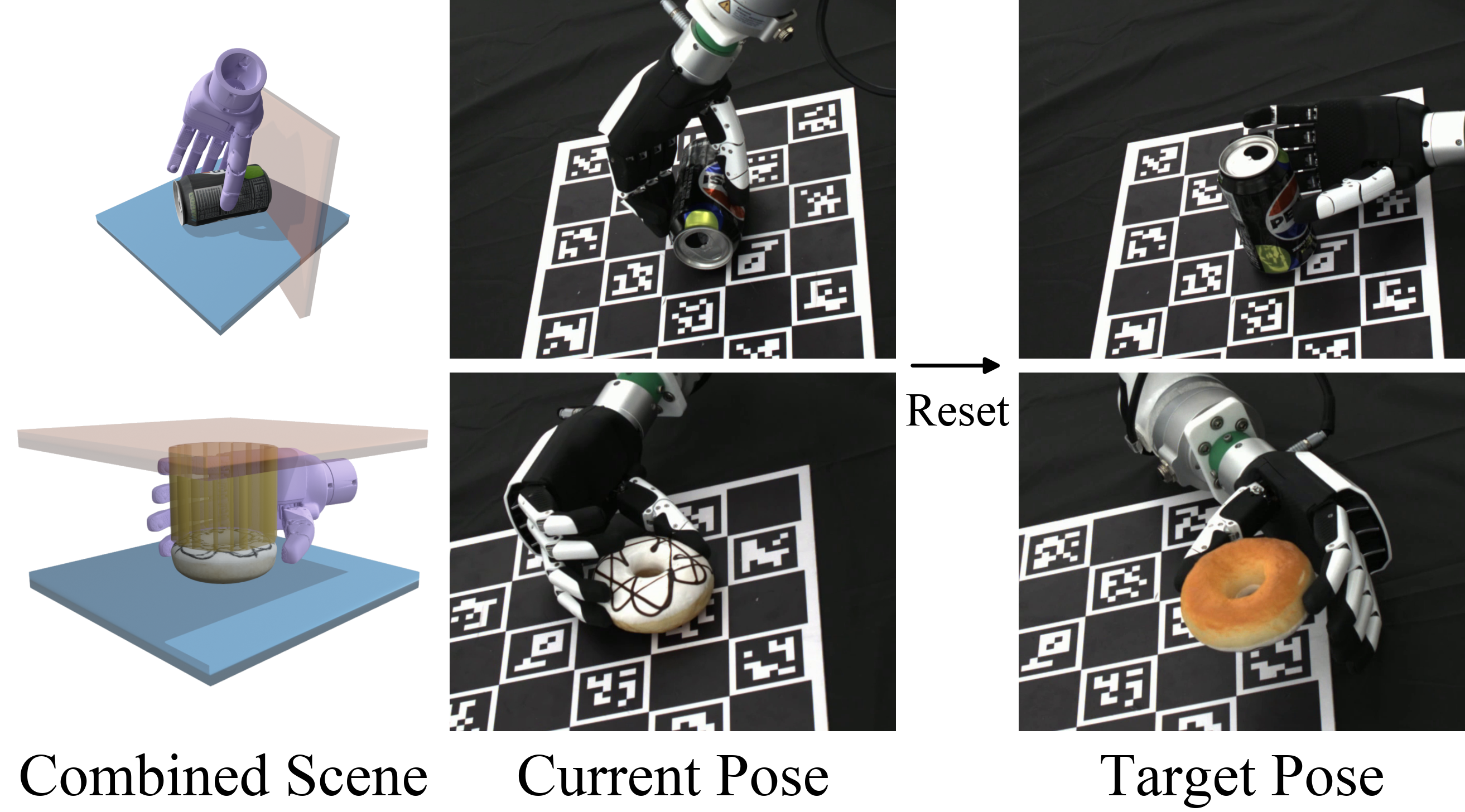}
\end{minipage}%
\hspace{0.005\linewidth}%
\begin{minipage}[t]{0.43\linewidth}
\centering
\includegraphics[height=4.2cm]
{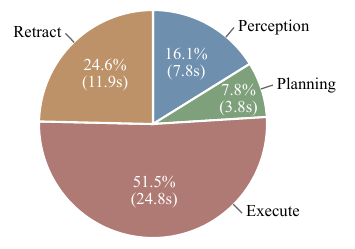}
\end{minipage}

\vspace{0pt}
\caption{
\textbf{Left: Reset examples.} The top row shows direct placement, where the robot carries the object to the target pose and releases it at the tabletop. The bottom row shows height-relaxed placement for a flat object, where virtual support pillars prevent finger intrusion into the object's descent region after release. In each row, the first panel shows the generated reset grasp, and the next two panels show the corresponding real-world pickup and placement motions.
\textbf{Right: Runtime analysis.} Runtime composition across 500 autonomous trials, showing that physical robot motion dominates the cycle.
}
\label{fig:reset_runtime}
\end{figure}

Each iteration of the collection loop begins by estimating the object's 6D pose from multi-view observations and identifying its current stable pose \(P_i \in \mathcal{P}\). AutoDex transforms the object-relative candidates in \(\mathcal{G}_{\mathrm{cand}}(P_i)\) into the robot frame and filters out candidates with no arm IK solution or with collisions. If no unattempted feasible candidate remains, AutoDex invokes the reset module in \cref{sec:method:reset} and returns to the beginning of the loop.

When an unattempted feasible candidate is available, AutoDex selects it for execution and plans a collision-free arm trajectory to its wrist pose, using an optimized open-hand configuration that maximizes clearance from the object and tabletop. The hand then moves through the candidate pre-grasp and grasp configurations, after which the robot lifts the object and holds for a fixed duration.
Each trial is labeled automatically from the captured multi-view stream. We run 6D object pose tracking across the stream and apply the lift-and-hold criterion: a trial is labeled successful if the tracked object remains at least 5\,cm above its initial height throughout the 3\,s hold phase, and as a failure otherwise.

After labeling, AutoDex saves the trial record, including the executed plan, multi-view recordings, tracked object poses, camera calibration, timing metadata, scene information, and the final success/failure label. Once recording ends, the robot places the object back on the table and, after it settles, returns to the start of the collection loop. The executed candidate is marked as attempted for the current stable pose to prevent repeated evaluation. This process continues until the candidate set is exhausted or the desired dataset size is reached.

\subsection{Reset}
\label{sec:method:reset}

To reset the object, AutoDex selects the least-seen reachable stable pose \(P_j\) with unattempted candidates remaining. The robot then reorients the object from the current pose \(P_i\) to \(P_j\). To generate a reset grasp, AutoDex combines the pickup constraints at \(P_i\) and the placement constraints at \(P_j\) into a single collision scene (\cref{fig:reset_runtime}, left). A grasp that is collision-free in this combined scene is geometrically compatible with both endpoints, allowing the same hand--object grasp to be used throughout the reset. For many pose pairs, the robot can then directly place the object: it grasps the object at \(P_i\), lifts and reorients it, places and releases it at \(P_j\), and retracts, as shown in the top row of \cref{fig:reset_runtime}.

For flat objects such as plates or trays, however, satisfying the pickup and placement constraints simultaneously can leave too little free space for the hand, even when a valid pickup grasp exists. AutoDex therefore relaxes the placement endpoint by allowing the object to be released from a height \(h\) above the target pose, as shown in the bottom row of \cref{fig:reset_runtime}. This relaxation introduces an additional constraint: after release, the object should descend to the target pose without contacting the fingers, since such contact can perturb the final object pose. During reset-grasp generation, virtual support pillars are added as collision geometry beneath the released object to exclude grasps whose fingers intrude into this descent region. We first try \(h=0\), corresponding to placement at the tabletop, and increase \(h\) only when no collision-free reset grasp is found. Reset grasps that successfully complete the physical transition are cached for each stable-pose pair and reused when the same transition is required again.

\begin{figure}[t]
\centering
\includegraphics[width=\linewidth]{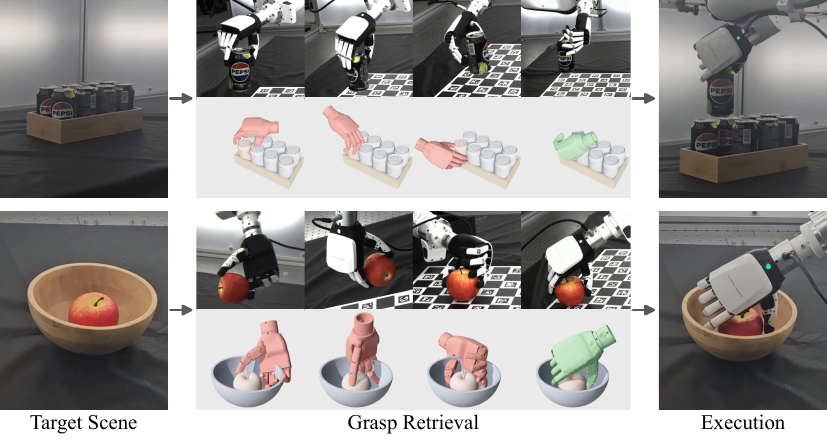}
\caption{
\textbf{Retrieval-based in-the-wild execution.} \textbf{Left:} Cluttered target scenes. \textbf{Middle:} Successful grasps retrieved from the AutoDex database (top), checked for collisions and kinematic reachability in the target scene (bottom). Red marks collision-rejected candidates, and green marks the feasible grasp selected for execution. \textbf{Right:} Real-world execution of the selected grasp.
}
\label{fig:placeholder}
\end{figure}

\subsection{AutoDex Database and Retrieval-Based In-the-Wild Execution}
\label{sec:method:algorithm}

We use AutoDex to collect a database of \numTraj{} physically executed and automatically labeled dexterous-grasp trials across the Allegro and Inspire hands. These comprise 1,912 successful and 1,681 failed trials. The Allegro and Inspire hands account for 2,226 and 1,367 trials, with success rates of 39.4\% and 75.7\%, respectively. The object set contains 100 household items, with over 80\% sourced from IKEA and the remainder drawn from common household goods; the objects span diverse geometries and materials, including plastic, metal, wood, silicone, paper, tape, and ceramic. Unlike a database of generated grasp candidates, the AutoDex database stores physical executions with success and failure outcomes. This allows real-world validation to be performed offline during data collection, while downstream use reduces to retrieving successful grasps and checking their feasibility in a new scene.

In our deployment setting, a robot observes the workspace with four calibrated RGB cameras and estimates the target object's 6D pose and nearby obstacle geometry. We use off-the-shelf tools to obtain object masks and an initial pose estimate~\cite{sam3,foundationpose}, refine the pose against multi-view silhouettes, and reconstruct surrounding geometry from depth estimates using Depth Anything 3~\cite{dav3}. Given the object pose and obstacle scene, we retrieve the successful grasps for the target object and transform each stored grasp into the workspace frame. A grasp is accepted only if a collision-free arm trajectory exists that reaches the corresponding wrist pose without intersecting the surrounding obstacles. The first grasp that passes this feasibility check is selected for execution. This retrieval procedure enables in-the-wild execution without policy training or additional real-world trial collection. \Cref{fig:placeholder} shows how validated grasps are retrieved and executed in constrained scenes.

\section{Experiments}
\label{sec:experiments}

We evaluate AutoDex as an autonomous system for collecting physically labeled real-world dexterous-grasp data. Our evaluation focuses on four aspects: (1) throughput and quality relative to teleoperation, (2) the effect of physical validation on the quality of a grasp database constructed from generator outputs, (3) the role of active reset in enabling efficient unattended collection across object poses, and (4) the effect of camera density on object-pose reliability during collection.
For evaluation, we use a representative 20-object subset drawn from our 100-object database. A grasp is considered successful if the object is lifted by 5 cm and held for 3 s.

\subsection{Data Collection Throughput, Quality, and Bottlenecks}
\label{sec:exp:throughput}

\begin{figure}[t]
\centering
\begin{minipage}[c]{0.33\linewidth}%
\centering
\footnotesize
\setlength{\tabcolsep}{2.5pt}
\begin{tabular}{lcc}
\toprule
 & Teleop. & AutoDex \\
\midrule
Wall (h) & 28.4 & \textbf{10.3} \\
Robot (h) & 8.3 & 10.1 \\
Human (h) & 9.3 & \textbf{0.2} \\
Interv. & always & \textbf{31} \\
\midrule
Success & \textbf{71.7\%} & 50.7\% \\
Jerk (rad/s$^3$) & 13.7 & \textbf{10.7} \\
Approach angle & $60^\circ$ & $\mathbf{82^\circ}$ \\
\bottomrule
\end{tabular}%
\end{minipage}\hspace{0.01\linewidth}%
\begin{minipage}[c]{0.66\linewidth}%
\centering
\includegraphics[width=\linewidth]{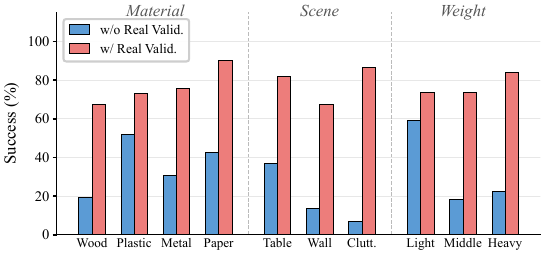}%
\end{minipage}
\vspace{-6pt}
\caption{
\textbf{Left: Comparison with teleoperation.} Collection time and human effort for 500 trajectories, together with grasp success, trajectory smoothness, and approach-direction diversity.
\textbf{Right: Effect of physical validation.} Grasps retrieved from the AutoDex-validated database achieve 76\% real-world success, compared with 34\% for grasps retrieved from the model-screened database, across 20 objects and 515 trials. The improvement is consistent across material, scene, and weight categories.
}
\label{fig:cumulative_traj_and_physical_validation}
\end{figure}

We compare AutoDex with teleoperation on collection time and human effort over 500 trajectories (\cref{fig:cumulative_traj_and_physical_validation}, left). Teleoperation requires 9.3\,h of human attention, comprising 8.3\,h of continuous operation and 1.0\,h of suit and setup preparation. In comparison, AutoDex requires an estimated 0.2\,h of human attention for object placement and recovery. Under a standardized eight-hour workday for teleoperation, collecting 500 trajectories takes an estimated 28.4\,h, compared with 10.3\,h for AutoDex. These results indicate that AutoDex reduces the need for sustained operator attention by allowing collection to proceed unattended between interventions.

We also compare grasp success, trajectory smoothness, and approach-direction diversity. Success rates are measured over the respective collection campaigns, with teleoperation achieving 71.7\% compared with 50.7\% for AutoDex. AutoDex produces smoother approach trajectories, with a mean joint jerk of 10.7\,rad/s$^3$ compared with 13.7\,rad/s$^3$ for teleoperation, measured after velocity smoothing. To evaluate approach-direction diversity, we express wrist approach axes in the object frame and compute the mean pairwise angle between grasps of the same object. Across 34 shared objects, AutoDex reaches $82^\circ$ compared with $60^\circ$ for teleoperation, indicating a broader range of approach directions.

We further analyze the runtime composition of AutoDex's autonomous collection loop in \cref{fig:reset_runtime} (right). Across 500 fully executed trials, the mean grasp-execution loop duration is 48.2 s. Robot execution accounts for 51.5\% of this loop (24.8 s), and retract accounts for 24.6\% (11.9 s), while perception and motion planning account for 16.1\% (7.8 s) and 7.8\% (3.8 s), respectively.
Most of the collection cycle is spent on robot execution and retraction, with perception and planning adding relatively little overhead.

\subsection{Impact of Physical Validation on Dataset Quality}
\label{sec:exp:physical_validation}

We evaluate whether physical validation improves the quality of a grasp database constructed from simulation-generated candidates. Experiments are conducted on the physical robot across three scene types, tabletop, wall, and cluttered, using a 20-object subset with 515 total trials.
We compare two databases constructed from the same generator output. The first is the model-screened database, which contains generated candidates screened only by the simulation sanity pre-filter, without real-world validation. The second is the AutoDex-validated database, which retains only candidates that were executed during autonomous collection and passed the lift-and-hold validity test. For each test object pose and scene, we apply the retrieval procedure in \cref{sec:method:algorithm}: candidates from the corresponding database are transformed into the workspace frame, filtered by collision-free arm planning, and executed on the robot.

\Cref{fig:cumulative_traj_and_physical_validation} (right) shows that the AutoDex-validated database achieves 76\% grasp success, compared with 34\% when using the model-screened database directly. The improvement holds across the evaluated material categories shown in \cref{fig:cumulative_traj_and_physical_validation}, as well as across scene types and weight bins, indicating that the gain is not due to a single object category or scene condition. The gap is largest in cluttered scenes, where grasps retrieved from the model-screened database frequently fail under real contact dynamics, while grasps retrieved from the AutoDex-validated database maintain 80-90\% success.
These results show that kinematic and collision feasibility alone are insufficient for constructing a reliable dexterous-grasp database. Physical validation substantially improves database quality by filtering out grasps that satisfy geometric constraints but fail under real-world contact effects such as slip, compliance, and multi-finger instability.

\subsection{Reset Strategy Evaluation}
\label{sec:exp:reset}

\begin{figure}[t]
\centering
\begin{minipage}[b]{0.44\linewidth}%
\centering
\includegraphics[width=\linewidth]{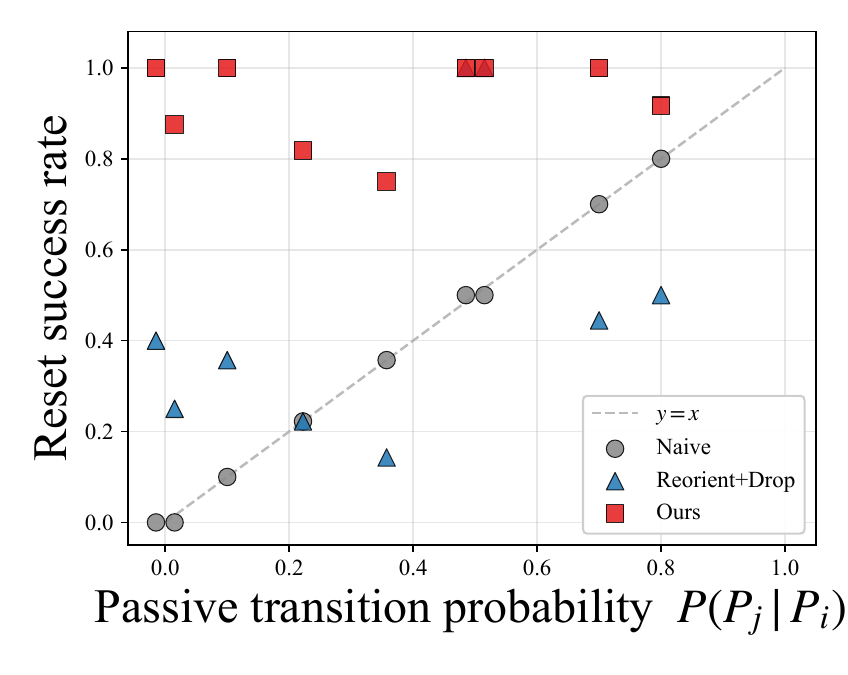}%
\end{minipage}\hspace{0.01\linewidth}%
\begin{minipage}[b]{0.55\linewidth}%
\centering
\includegraphics[width=\linewidth]{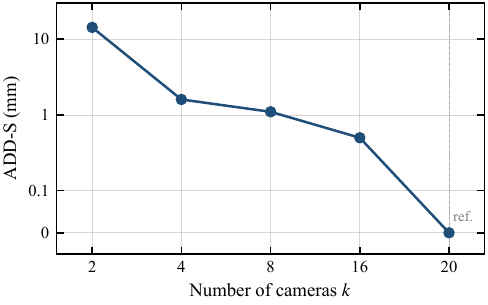}%
\end{minipage}
\vspace{-18pt}
\caption{
\textbf{Left: Reset strategy comparison.} Reset success versus passive transition probability \(P(P_j \mid P_i)\). Naive Drop follows \(y=x\) by construction, while Stable Reorient Placement maintains high success even for transitions rarely reached by passive settling.
\textbf{Right: Pose self-consistency relative to the 20-camera reference as a function of camera count.} Mean ADD-S between the full 20-camera reference pose and \(k\)-camera subset poses, over 51 trials and 5 objects.
}
\label{fig:reset_eval_and_cameras_vs_success}
\end{figure}

We compare our reset method against two passive-drop baselines. \textit{Naive Drop} lifts the object straight up and releases it, allowing it to settle under gravity. \textit{Reorient + Passive Release} first reorients the object in the gripper toward a target pose \(P_j\), then releases it above the table. \textit{Stable Reorient Placement}, our method, places the object onto the table in the target pose before release, removing the dependence on passive settling.

Pose transitions are not equally difficult. Some target poses naturally appear after passive dropping, while others almost never occur without deliberate placement. We therefore evaluate reset success as a function of passive transition probability. We count a reset as successful when the object reaches the target stable pose \(P_j\). For each object, we estimate this probability by repeatedly dropping it from tabletop pose \(P_i\) and recording the resulting pose \(P_j\), yielding \(P(P_j \mid P_i)\). Low values of \(P(P_j \mid P_i)\) correspond to hard reset transitions, because passive settling rarely produces the desired target pose.

\Cref{fig:reset_eval_and_cameras_vs_success} (left) reports reset success against \(P(P_j \mid P_i)\). By construction, \textit{Naive Drop} tracks \(y=x\): its success rate equals the passive transition probability. \textit{Reorient + Passive Release} improves over \textit{Naive Drop} in some cases but remains close to the passive baseline because the final pose is still determined by uncontrolled settling after release. \textit{Ours} maintains consistently high success across the full difficulty range, including transitions with \(P(P_j \mid P_i) \approx 0\) that are essentially unreachable by passive strategies. \textit{Reorient + Passive Release} also exhibits a failure mode incompatible with unattended operation: it ejects the object from the workspace in 5.3\% of attempts, requiring human recovery, whereas our method causes no such failures. These results show that explicit reset placement is critical for reliable unattended dexterous-grasp collection.

\subsection{Impact of Camera Density on Pose Reliability}
\label{sec:exp:cameras}

We study how perception quality scales with the number of cameras used during silhouette refinement. For each capture, we treat the pose refined from all 20 cameras as the reference and re-refine the object pose using a random subset of \(k\) cameras. \Cref{fig:reset_eval_and_cameras_vs_success} (right) reports the mean ADD-S~\cite{hinterstoisser2012model} between the reference pose and the subset pose for \(k \in \{2,4,8,16\}\).
Mean ADD-S decreases from 14.3 mm at \(k=2\) to 0.5 mm at \(k=16\), with the largest improvement occurring between \(k=2\) and \(k=4\). The high error at \(k=2\) is driven by occasional catastrophic pose failures on textureless or thin objects; these tail failures are largely eliminated by \(k=8\). Because candidate selection, collision checking, and grasp execution all depend on the estimated object pose, this result indicates that dense multi-view redundancy improves pose reliability for unattended collection.

\section{Discussion and Limitations}
\label{sec:limitations}

AutoDex shifts the operator's role from controlling individual grasp attempts to initially placing the object in the workcell and resolving failures that the system cannot recover from automatically. Integrating execution, collision monitoring, labeling, and reset allows collection to proceed without continuous operator attention. The use of a grasp generator also makes the candidate set an explicit input to collection, allowing users to specify grasp configurations and scene constraints rather than relying on an operator to select each attempt. Users can therefore control which candidates are tested, although collection remains limited to the supplied candidate set. The lift-and-hold criterion assesses grasp stability but does not establish suitability for subsequent tasks such as tool use or handover. While the 20-camera workcell supports reliable perception during collection, its hardware requirements may limit accessibility. Because the collection pipeline also operates with as few as two cameras, users can collect their own data with a smaller setup; alternatively, they can use the released grasp database directly. Our retrieval experiments demonstrate how validated grasps can be reused under new object placements and obstacle configurations, within the scope of the objects and hands represented in the database.

\acknowledgments{This work was supported by RLWRLD, Institute of Engineering Research at Seoul National University, KIST Institutional Program (26Z9020), NRF grant funded by the Korean government (MSIT) [No. RS-2025-25396144, and No. RS-2026-25540546], and IITP grant funded by the Korea government (MSIT) [No. RS-2024-00439854, No. RS-2025-25442338, and No. RS-2021-II211343]. H. Joo is the corresponding author.}

\bibliography{references}

\begin{thebibliography}{59}
\providecommand{\natexlab}[1]{#1}
\providecommand{\url}[1]{\texttt{#1}}
\expandafter\ifx\csname urlstyle\endcsname\relax
  \providecommand{\doi}[1]{doi: #1}\else
  \providecommand{\doi}{doi: \begingroup \urlstyle{rm}\Url}\fi

\bibitem[Liu et~al.(2024)Liu, Yang, Wang, Wu, Wang, Yao, Schwertfeger, Yang,
  Wang, Yu, He, and Ma]{liu2024realdex}
Y.~Liu, Y.~Yang, Y.~Wang, X.~Wu, J.~Wang, Y.~Yao, S.~Schwertfeger, S.~Yang,
  W.~Wang, J.~Yu, X.~He, and Y.~Ma.
\newblock {RealDex}: Towards human-like grasping for robotic dexterous hand.
\newblock In \emph{IJCAI}, 2024.

\bibitem[Wang et~al.(2024)Wang, Shi, Wang, Zhang, Fei-Fei, and
  Liu]{wang2024dexcap}
C.~Wang, H.~Shi, W.~Wang, R.~Zhang, L.~Fei-Fei, and C.~K. Liu.
\newblock {DexCap}: Scalable and portable mocap data collection system for
  dexterous manipulation.
\newblock In \emph{RSS}, 2024.

\bibitem[Chen et~al.(2022)Chen, Van~Wyk, Chao, Yang, Mousavian, Gupta, and
  Fox]{chen2022dextransfer}
Z.~Chen, K.~Van~Wyk, Y.-W. Chao, W.~Yang, A.~Mousavian, A.~Gupta, and D.~Fox.
\newblock {DexTransfer}: Real world multi-fingered dexterous grasping with
  minimal human demonstrations.
\newblock \emph{arXiv preprint arXiv:2209.14284}, 2022.

\bibitem[Zhang et~al.(2024)Zhang, Liu, Li, Yu, Geng, Ding, Chen, and
  Wang]{zhang2024dexgraspnet20learninggenerative}
J.~Zhang, H.~Liu, D.~Li, X.~Yu, H.~Geng, Y.~Ding, J.~Chen, and H.~Wang.
\newblock {DexGraspNet} 2.0: Learning generative dexterous grasping in
  large-scale synthetic cluttered scenes.
\newblock In \emph{CoRL}, 2024.

\bibitem[Bicchi and Kumar(2000)]{bicchi2000grasping}
A.~Bicchi and V.~Kumar.
\newblock Robotic grasping and contact: A review.
\newblock In \emph{ICRA}, 2000.

\bibitem[Qin et~al.(2023)Qin, Yang, Huang, Van~Wyk, Su, Wang, Chao, and
  Fox]{qin2023anyteleop}
Y.~Qin, W.~Yang, B.~Huang, K.~Van~Wyk, H.~Su, X.~Wang, Y.-W. Chao, and D.~Fox.
\newblock {AnyTeleop}: A general vision-based dexterous robot arm-hand
  teleoperation system.
\newblock In \emph{RSS}, 2023.

\bibitem[Wang et~al.(2023)Wang, Zhang, Chen, Xu, Li, Liu, and
  Wang]{wang2022dexgraspnet}
R.~Wang, J.~Zhang, J.~Chen, Y.~Xu, P.~Li, T.~Liu, and H.~Wang.
\newblock {DexGraspNet}: A large-scale robotic dexterous grasp dataset for
  general objects based on simulation.
\newblock In \emph{ICRA}, 2023.

\bibitem[Chen et~al.(2025)Chen, Ke, and Wang]{chen2024bodex}
J.~Chen, Y.~Ke, and H.~Wang.
\newblock {BODex}: Scalable and efficient robotic dexterous grasp synthesis
  using bilevel optimization.
\newblock In \emph{ICRA}, 2025.

\bibitem[Turpin et~al.(2023)Turpin, Zhong, Zhang, Zhu, Liu, Singh, Heiden,
  Macklin, Tsogkas, Dickinson, et~al.]{turpin2023fast}
D.~Turpin, T.~Zhong, S.~Zhang, G.~Zhu, J.~Liu, R.~Singh, E.~Heiden, M.~Macklin,
  S.~Tsogkas, S.~Dickinson, et~al.
\newblock {Fast-Grasp'D}: Dexterous multi-finger grasp generation through
  differentiable simulation.
\newblock In \emph{ICRA}, 2023.

\bibitem[Huang et~al.(2025)Huang, Zhang, Li, Zhao, Li, Fang, Xia, and
  He]{huang2025dexgrasp}
D.~Huang, T.~Zhang, Y.~Li, L.~Zhao, J.~Li, Z.~Fang, C.~Xia, and X.~He.
\newblock Dexterous grasping with real-world robotic reinforcement learning.
\newblock \emph{arXiv preprint arXiv:2503.04014}, 2025.

\bibitem[Lim et~al.(2026)Lim, Ha, Cha, Cho, Choi, Jeon, Kim, Kim, and
  Joo]{lim2026hrdexdb}
J.~Lim, T.~Ha, S.~Cha, K.~Cho, M.~Choi, S.~Jeon, J.~Kim, B.~Kim, and H.~Joo.
\newblock {HRDexDB}: A {4D} dexterous grasping dataset across human and
  multiple robot embodiments.
\newblock \emph{arXiv preprint arXiv:2604.14944}, 2026.

\bibitem[Wang et~al.(2025)Wang, Ye, Xiao, Zhong, Tao, Yu, Liu, Yu, and
  Ma]{wang2025dexh2r}
Y.~Wang, J.~Ye, C.~Xiao, Y.~Zhong, H.~Tao, H.~Yu, Y.~Liu, J.~Yu, and Y.~Ma.
\newblock {DexH2R}: A benchmark for dynamic dexterous grasping in
  human-to-robot handover.
\newblock In \emph{ICCV}, 2025.

\bibitem[Li et~al.(2026)Li, Wu, Zhang, Li, Gan, Ouyang, and
  Wang]{li2026bimanualgrasp}
Z.~Li, M.~Wu, J.~Zhang, H.~Li, Y.~Gan, D.~Ouyang, and N.~Wang.
\newblock Real-world cooperative bimanual dexterous grasp of large objects from
  single-view observations.
\newblock In \emph{IROS}, 2026.

\bibitem[Handa et~al.(2020)Handa, Van~Wyk, Yang, Liang, Chao, Wan, Birchfield,
  Ratliff, and Fox]{handa2020dexpilot}
A.~Handa, K.~Van~Wyk, W.~Yang, J.~Liang, Y.-W. Chao, Q.~Wan, S.~Birchfield,
  N.~Ratliff, and D.~Fox.
\newblock {DexPilot}: Vision-based teleoperation of dexterous robotic hand-arm
  system.
\newblock In \emph{ICRA}, 2020.

\bibitem[Cheng et~al.(2024)Cheng, Li, Yang, Yang, and
  Wang]{cheng2024opentelevision}
X.~Cheng, J.~Li, S.~Yang, G.~Yang, and X.~Wang.
\newblock {Open-TeleVision}: Teleoperation with immersive active visual
  feedback.
\newblock In \emph{CoRL}, 2024.

\bibitem[Ding et~al.(2025)Ding, Qin, Zhu, Jia, Yang, Yang, Qi, and
  Wang]{ding2024bunnyvisionpro}
R.~Ding, Y.~Qin, J.~Zhu, C.~Jia, S.~Yang, R.~Yang, X.~Qi, and X.~Wang.
\newblock {Bunny-VisionPro}: Real-time bimanual dexterous teleoperation for
  imitation learning.
\newblock In \emph{IROS}, 2025.

\bibitem[Qin et~al.(2022)Qin, Wu, Liu, Jiang, Yang, Fu, and Wang]{qin2022dexmv}
Y.~Qin, Y.-H. Wu, S.~Liu, H.~Jiang, R.~Yang, Y.~Fu, and X.~Wang.
\newblock {DexMV}: Imitation learning for dexterous manipulation from human
  videos.
\newblock In \emph{ECCV}, 2022.

\bibitem[Shaw et~al.(2022)Shaw, Bahl, and Pathak]{shaw2022videodex}
K.~Shaw, S.~Bahl, and D.~Pathak.
\newblock {VideoDex}: Learning dexterity from internet videos.
\newblock In \emph{CoRL}, 2022.

\bibitem[Tao et~al.(2025)Tao, Srirama, Liu, Shaw, and Pathak]{tao2025dexwild}
T.~Tao, M.~K. Srirama, J.~J. Liu, K.~Shaw, and D.~Pathak.
\newblock {DexWild}: Dexterous human interactions for in-the-wild robot
  policies.
\newblock In \emph{RSS}, 2025.

\bibitem[Chi et~al.(2024)Chi, Xu, Pan, Cousineau, Burchfiel, Feng, Tedrake, and
  Song]{chi2024umi}
C.~Chi, Z.~Xu, C.~Pan, E.~Cousineau, B.~Burchfiel, S.~Feng, R.~Tedrake, and
  S.~Song.
\newblock Universal manipulation interface: In-the-wild robot teaching without
  in-the-wild robots.
\newblock In \emph{RSS}, 2024.

\bibitem[Fang et~al.(2024)Fang, Fang, Wang, Ren, Chen, Zhang, Wang, and
  Lu]{fang2024airexo}
H.~Fang, H.-S. Fang, Y.~Wang, J.~Ren, J.~Chen, R.~Zhang, W.~Wang, and C.~Lu.
\newblock {AirExo}: Low-cost exoskeletons for learning whole-arm manipulation
  in the wild.
\newblock In \emph{ICRA}, 2024.

\bibitem[Xu et~al.(2025)Xu, Zhang, Hou, Xu, Fan, Veloso, and
  Song]{xu2025dexumi}
M.~Xu, H.~Zhang, Y.~Hou, Z.~Xu, L.~Fan, M.~Veloso, and S.~Song.
\newblock {DexUMI}: Using human hand as the universal manipulation interface
  for dexterous manipulation.
\newblock In \emph{CoRL}, 2025.

\bibitem[Park et~al.(2025)Park, Bhatia, Ankile, and Agrawal]{park2025dart}
Y.~Park, J.~S. Bhatia, L.~Ankile, and P.~Agrawal.
\newblock {DART}: Dexterous augmented reality teleoperation platform for
  large-scale robot data collection in simulation.
\newblock In \emph{ICRA}, 2025.

\bibitem[Chen et~al.(2025)Chen, Wang, Nguyen, Fei-Fei, and Liu]{chen2025arcap}
S.~Chen, C.~Wang, K.~Nguyen, L.~Fei-Fei, and C.~K. Liu.
\newblock {ARCap}: Collecting high-quality human demonstrations for robot
  learning with augmented reality feedback.
\newblock In \emph{ICRA}, 2025.

\bibitem[Miller and Allen(2004)]{miller2004graspit}
A.~T. Miller and P.~K. Allen.
\newblock {GraspIt!} {A} versatile simulator for robotic grasping.
\newblock \emph{IEEE RAM}, 2004.

\bibitem[Liu et~al.(2022)Liu, Liu, Jiao, Zhu, and Zhu]{Liu_2022}
T.~Liu, Z.~Liu, Z.~Jiao, Y.~Zhu, and S.-C. Zhu.
\newblock Synthesizing diverse and physically stable grasps with arbitrary hand
  structures using differentiable force closure estimator.
\newblock \emph{IEEE RA-L}, 2022.

\bibitem[Turpin et~al.(2022)Turpin, Wang, Heiden, Chen, Macklin, Tsogkas,
  Dickinson, and Garg]{turpin2022graspd}
D.~Turpin, L.~Wang, E.~Heiden, Y.-C. Chen, M.~Macklin, S.~Tsogkas,
  S.~Dickinson, and A.~Garg.
\newblock {Grasp'D}: Differentiable contact-rich grasp synthesis for
  multi-fingered hands.
\newblock In \emph{ECCV}, 2022.

\bibitem[Chen et~al.(2025)Chen, Ke, Peng, and Wang]{chen2025dexonomy}
J.~Chen, Y.~Ke, L.~Peng, and H.~Wang.
\newblock {Dexonomy}: Synthesizing all dexterous grasp types in a grasp
  taxonomy.
\newblock In \emph{RSS}, 2025.

\bibitem[Xu et~al.(2023)Xu, Wan, Zhang, Liu, Shan, Shen, Wang, Geng, Weng,
  Chen, et~al.]{xu2023unidexgrasp}
Y.~Xu, W.~Wan, J.~Zhang, H.~Liu, Z.~Shan, H.~Shen, R.~Wang, H.~Geng, Y.~Weng,
  J.~Chen, et~al.
\newblock {UniDexGrasp}: Universal robotic dexterous grasping via learning
  diverse proposal generation and goal-conditioned policy.
\newblock In \emph{CVPR}, 2023.

\bibitem[Feng et~al.(2025)Feng, Feng, Chen, Triebel, and
  Knoll]{feng2025ffhflow}
Q.~Feng, J.~Feng, Z.~Chen, R.~Triebel, and A.~Knoll.
\newblock {FFHFlow}: Diverse and uncertainty-aware dexterous grasp generation
  via flow variational inference.
\newblock In \emph{CoRL}, 2025.

\bibitem[Christen et~al.(2022)Christen, Kocabas, Aksan, Hwangbo, Song, and
  Hilliges]{christen2022d}
S.~Christen, M.~Kocabas, E.~Aksan, J.~Hwangbo, J.~Song, and O.~Hilliges.
\newblock {D-Grasp}: Physically plausible dynamic grasp synthesis for
  hand-object interactions.
\newblock In \emph{CVPR}, 2022.

\bibitem[Qin et~al.(2022)Qin, Huang, Yin, Su, and Wang]{qin2022dexpoint}
Y.~Qin, B.~Huang, Z.-H. Yin, H.~Su, and X.~Wang.
\newblock {DexPoint}: Generalizable point cloud reinforcement learning for
  sim-to-real dexterous manipulation.
\newblock In \emph{CoRL}, 2022.

\bibitem[Wan et~al.(2023)Wan, Geng, Liu, Shan, Yang, Yi, and
  Wang]{wan2023unidexgrasp++}
W.~Wan, H.~Geng, Y.~Liu, Z.~Shan, Y.~Yang, L.~Yi, and H.~Wang.
\newblock {UniDexGrasp++}: Improving dexterous grasping policy learning via
  geometry-aware curriculum and iterative generalist-specialist learning.
\newblock In \emph{ICCV}, 2023.

\bibitem[Zhang et~al.(2025)Zhang, Wu, Huang, Christen, and
  Song]{zhang2025robustdexgrasp}
H.~Zhang, Z.~Wu, L.~Huang, S.~Christen, and J.~Song.
\newblock {RobustDexGrasp}: Robust dexterous grasping of general objects.
\newblock In \emph{CoRL}, 2025.

\bibitem[Lum et~al.(2024)Lum, Li, Culbertson, Srinivasan, Ames, Schwager, and
  Bohg]{lum2025getagrip}
T.~G.~W. Lum, A.~H. Li, P.~Culbertson, K.~Srinivasan, A.~Ames, M.~Schwager, and
  J.~Bohg.
\newblock {Get a Grip}: Multi-finger grasp evaluation at scale enables robust
  sim-to-real transfer.
\newblock In \emph{CoRL}, 2024.

\bibitem[Ye et~al.(2025)Ye, Wang, Yuan, Yang, Li, Zhu, Qin, Zou, and
  Wang]{ye2025dex1b}
J.~Ye, K.~Wang, C.~Yuan, R.~Yang, Y.~Li, J.~Zhu, Y.~Qin, X.~Zou, and X.~Wang.
\newblock {Dex1B}: Learning with {1B} demonstrations for dexterous
  manipulation.
\newblock In \emph{RSS}, 2025.

\bibitem[Mahler et~al.(2017)Mahler, Liang, Niyaz, Laskey, Doan, Liu,
  Aparicio~Ojea, and Goldberg]{mahler2017dexnet2}
J.~Mahler, J.~Liang, S.~Niyaz, M.~Laskey, R.~Doan, X.~Liu, J.~Aparicio~Ojea,
  and K.~Goldberg.
\newblock {Dex-Net 2.0}: Deep learning to plan robust grasps with synthetic
  point clouds and analytic grasp metrics.
\newblock In \emph{RSS}, 2017.

\bibitem[Mousavian et~al.(2019)Mousavian, Eppner, and
  Fox]{mousavian2019graspnet}
A.~Mousavian, C.~Eppner, and D.~Fox.
\newblock {6-DOF GraspNet}: Variational grasp generation for object
  manipulation.
\newblock In \emph{ICCV}, 2019.

\bibitem[Horak and Trinkle(2019)]{horak2019contact}
P.~C. Horak and J.~C. Trinkle.
\newblock On the similarities and differences among contact models in robot
  simulation.
\newblock \emph{IEEE RA-L}, 2019.

\bibitem[Rubert et~al.(2019)Rubert, Kappler, Bohg, and
  Morales]{rubert2019predicting}
C.~Rubert, D.~Kappler, J.~Bohg, and A.~Morales.
\newblock Predicting grasp success in the real world -- a study of quality
  metrics and human assessment.
\newblock \emph{RAS}, 2019.

\bibitem[Le~Lidec et~al.(2024)Le~Lidec, Jallet, Montaut, Laptev, Schmid, and
  Carpentier]{lelidec2024contact}
Q.~Le~Lidec, W.~Jallet, L.~Montaut, I.~Laptev, C.~Schmid, and J.~Carpentier.
\newblock Contact models in robotics: A comparative analysis.
\newblock \emph{IEEE T-RO}, 2024.

\bibitem[Pinto and Gupta(2016)]{pinto2016supersizing}
L.~Pinto and A.~Gupta.
\newblock Supersizing self-supervision: Learning to grasp from {50K} tries and
  700 robot hours.
\newblock In \emph{ICRA}, 2016.

\bibitem[Gupta et~al.(2018)Gupta, Murali, Gandhi, and Pinto]{gupta2018homes}
A.~Gupta, A.~Murali, D.~Gandhi, and L.~Pinto.
\newblock Robot learning in homes: Improving generalization and reducing
  dataset bias.
\newblock In \emph{NeurIPS}, 2018.

\bibitem[Levine et~al.(2018)Levine, Pastor, Krizhevsky, Ibarz, and
  Quillen]{levine2018handeye}
S.~Levine, P.~Pastor, A.~Krizhevsky, J.~Ibarz, and D.~Quillen.
\newblock Learning hand-eye coordination for robotic grasping with deep
  learning and large-scale data collection.
\newblock \emph{IJRR}, 2018.

\bibitem[Kalashnikov et~al.(2018)Kalashnikov, Irpan, Pastor, Ibarz, Herzog,
  Jang, Quillen, Holly, Kalakrishnan, Vanhoucke, and
  Levine]{kalashnikov2018qtopt}
D.~Kalashnikov, A.~Irpan, P.~Pastor, J.~Ibarz, A.~Herzog, E.~Jang, D.~Quillen,
  E.~Holly, M.~Kalakrishnan, V.~Vanhoucke, and S.~Levine.
\newblock Scalable deep reinforcement learning for vision-based robotic
  manipulation.
\newblock In \emph{CoRL}, 2018.

\bibitem[Kalashnikov et~al.(2021)Kalashnikov, Varley, Chebotar, Swanson,
  Jonschkowski, Finn, Levine, and Hausman]{kalashnikov2021mtopt}
D.~Kalashnikov, J.~Varley, Y.~Chebotar, B.~Swanson, R.~Jonschkowski, C.~Finn,
  S.~Levine, and K.~Hausman.
\newblock Scaling up multi-task robotic reinforcement learning.
\newblock In \emph{CoRL}, 2021.

\bibitem[Sharma et~al.(2023)Sharma, Ahmed, Ahmad, and
  Finn]{sharma2023selfimproving}
A.~Sharma, A.~M. Ahmed, R.~Ahmad, and C.~Finn.
\newblock Self-improving robots: End-to-end autonomous visuomotor reinforcement
  learning.
\newblock In \emph{CoRL}, 2023.

\bibitem[Zhou et~al.(2024)Zhou, Atreya, Lee, Walke, Mees, and
  Levine]{zhou2024soar}
Z.~Zhou, P.~Atreya, A.~Lee, H.~Walke, O.~Mees, and S.~Levine.
\newblock Autonomous improvement of instruction following skills via foundation
  models.
\newblock In \emph{CoRL}, 2024.

\bibitem[Liu et~al.(2023)Liu, Nasiriany, Zhang, Bao, and Zhu]{liu2023sirius}
H.~Liu, S.~Nasiriany, L.~Zhang, Z.~Bao, and Y.~Zhu.
\newblock Robot learning on the job: Human-in-the-loop autonomy and learning
  during deployment.
\newblock In \emph{RSS}, 2023.

\bibitem[Ahn et~al.(2024)Ahn, Dwibedi, Finn, {Gonzalez Arenas}, Gopalakrishnan,
  Hausman, Ichter, Irpan, Joshi, Julian, et~al.]{ahn2024autort}
M.~Ahn, D.~Dwibedi, C.~Finn, M.~{Gonzalez Arenas}, K.~Gopalakrishnan,
  K.~Hausman, B.~Ichter, A.~Irpan, N.~Joshi, R.~Julian, et~al.
\newblock {AutoRT}: Embodied foundation models for large scale orchestration of
  robotic agents.
\newblock \emph{arXiv preprint arXiv:2401.12963}, 2024.

\bibitem[Yang et~al.(2024)Yang, Mark, Vu, Sharma, Bohg, and
  Finn]{yang2023robofume}
J.~Yang, M.~S. Mark, B.~Vu, A.~Sharma, J.~Bohg, and C.~Finn.
\newblock Robot fine-tuning made easy: Pre-training rewards and policies for
  autonomous real-world reinforcement learning.
\newblock In \emph{ICRA}, 2024.

\bibitem[Eysenbach et~al.(2018)Eysenbach, Gu, Ibarz, and
  Levine]{eysenbach2018leave}
B.~Eysenbach, S.~Gu, J.~Ibarz, and S.~Levine.
\newblock Leave no trace: Learning to reset for safe and autonomous
  reinforcement learning.
\newblock In \emph{ICLR}, 2018.

\bibitem[Zhu et~al.(2020)Zhu, Yu, Gupta, Shah, Hartikainen, Singh, Kumar, and
  Levine]{zhu2020ingredients}
H.~Zhu, J.~Yu, A.~Gupta, D.~Shah, K.~Hartikainen, A.~Singh, V.~Kumar, and
  S.~Levine.
\newblock The ingredients of real-world robotic reinforcement learning.
\newblock In \emph{ICLR}, 2020.

\bibitem[Gupta et~al.(2021)Gupta, Yu, Zhao, Kumar, Rovinsky, Xu, Devlin, and
  Levine]{gupta2021resetfree}
A.~Gupta, J.~Yu, T.~Z. Zhao, V.~Kumar, A.~Rovinsky, K.~Xu, T.~Devlin, and
  S.~Levine.
\newblock Reset-free reinforcement learning via multi-task learning: Learning
  dexterous manipulation behaviors without human intervention.
\newblock In \emph{ICRA}, 2021.

\bibitem[Du et~al.(2023)Du, Konyushkova, Denil, Raju, Landon, Hill, de~Freitas,
  and Cabi]{du2023success}
Y.~Du, K.~Konyushkova, M.~Denil, A.~Raju, J.~Landon, F.~Hill, N.~de~Freitas,
  and S.~Cabi.
\newblock Vision-language models as success detectors.
\newblock In \emph{CoLLAs}, 2023.

\bibitem[Carion et~al.(2026)Carion, Gustafson, Hu, Debnath, Hu, Suris, Ryali,
  Alwala, Khedr, Huang, et~al.]{sam3}
N.~Carion, L.~Gustafson, Y.-T. Hu, S.~Debnath, R.~Hu, D.~Suris, C.~Ryali, K.~V.
  Alwala, H.~Khedr, A.~Huang, et~al.
\newblock {SAM 3}: Segment anything with concepts.
\newblock In \emph{ICLR}, 2026.

\bibitem[Wen et~al.(2024)Wen, Yang, Kautz, and Birchfield]{foundationpose}
B.~Wen, W.~Yang, J.~Kautz, and S.~Birchfield.
\newblock {FoundationPose}: Unified {6D} pose estimation and tracking of novel
  objects.
\newblock In \emph{CVPR}, 2024.

\bibitem[Lin et~al.(2026)Lin, Chen, Liew, Chen, Li, Zhao, Peng, Guo, Zhou, Shi,
  Feng, and Kang]{dav3}
H.~Lin, S.~Chen, J.~H. Liew, D.~Y. Chen, Z.~Li, Y.~Zhao, S.~Peng, H.~Guo,
  X.~Zhou, G.~Shi, J.~Feng, and B.~Kang.
\newblock {Depth Anything 3}: Recovering the visual space from any views.
\newblock In \emph{ICLR}, 2026.

\bibitem[Hinterstoisser et~al.(2012)Hinterstoisser, Lepetit, Ilic, Holzer,
  Bradski, Konolige, and Navab]{hinterstoisser2012model}
S.~Hinterstoisser, V.~Lepetit, S.~Ilic, S.~Holzer, G.~Bradski, K.~Konolige, and
  N.~Navab.
\newblock Model based training, detection and pose estimation of texture-less
  {3D} objects in heavily cluttered scenes.
\newblock In \emph{ACCV}, 2012.

\end{thebibliography}

\end{document}